\pdfoutput=1
\documentclass[twoside,11pt]{article}

\usepackage[accepted]{melba}

\usepackage{mwe} 

\usepackage{amsmath,amsfonts}

\usepackage{graphicx}
\usepackage{amssymb, amsfonts, pifont}
\usepackage{booktabs}
\usepackage{breqn}
\usepackage{algorithm,algorithmic}
\usepackage{textcomp}
\usepackage{comment}
\usepackage{todonotes}
\usepackage{multicol, multirow}
\newcommand{\cmark}{\ding{51}}%
\newcommand{\xmark}{\ding{55}}%
\usepackage{caption}
\usepackage{subcaption}
\usepackage{tabularx}
\usepackage{ragged2e}
\usepackage{booktabs}
\usepackage{makecell}
\usepackage{hyperref}

\melbaid{2024:015}  
\doi{https://doi.org/10.59275/j.melba.2024-258b}
\melbaauthors{Sinhamahapatra et al}  
\volume{2}
\firstpageno{977}  
\melbayear{2024}  
\datesubmitted{12/2023}  
\datepublished{07/2024}  

\melbaspecialissue{Medical Imaging with Deep Learning (MIDL) 2020}
\melbaspecialissueeditors{Marleen de Bruijne, Tal Arbel, Ismail Ben Ayed, Hervé Lombaert}

\ShortHeadings{Interpretability of DL-assisted VCF Grading}{Sinhamahapatra et al.}

\title{Enhancing Interpretability of Vertebrae Fracture Grading \\ 
using Human-interpretable Prototypes}

\author{\firstname Poulami \surname Sinhamahapatra \orcid{0000-0002-3873-9623} \email poulami.sinhamahapatra@iks.fraunhofer.de \\ 
 \addr Fraunhofer IKS \and Technical University of Munich, Germany
 \AND
 \firstname Suprosanna \surname Shit \orcid{0000-0003-4435-7207} \email suprosanna.shit@tum.de \\
 \addr Technical University of Munich, Germany
 \AND
 \firstname Anjany \surname Sekuboyina \orcid{0000-0002-5601-284X} \email anjany.sekuboyina@tum.de \\
 \addr Technical University of Munich, Germany
 \AND
 \firstname Malek El \surname Husseini \orcid{0000-0002-2952-3708} \email malek.husseini@tum.de \\
 \addr Technical University of Munich, Germany
 \AND
 \firstname David \surname Schinz \orcid{0000-0003-3734-1135} \email david.schinz@tum.de \\
 \addr Technical University of Munich, Germany
 \AND
 \firstname Nicolas \surname Lenhart \email nicolas.lenhart@tum.de \\
 \addr Technical University of Munich, Germany
 \AND
 \firstname Bjoern \surname Menze \orcid{0000-0003-4136-5690} \email bjoern.menze@uzh.ch \\
 \addr University of Zurich, Switzerland
 \AND
 \firstname Jan \surname Kirschke \orcid{0000-0002-7557-0003} \email jan.kirschke@tum.de \\
 \addr Technical University of Munich, Germany
 \AND
 \firstname Karsten \surname Roscher \orcid{0000-0002-9458-104X} \email karsten.roscher@iks.fraunhofer.de \\
 \addr Fraunhofer IKS, Munich, Germany
 \AND
 \firstname Stephan \surname Guennemann \orcid{0000-0001-7772-5059} \email guennemann@in.tum.de \\
 \addr Technical University of Munich, Germany
}

\begin{document}

\maketitle

\begin{abstract}
	Vertebral fracture grading classifies the severity of vertebral fractures, which is a challenging task in medical imaging and has recently attracted Deep Learning (DL) models. Only a few works attempted to make such models human-interpretable despite the need for transparency and trustworthiness in critical use cases like DL-assisted medical diagnosis. Moreover, such models either rely on post-hoc methods or additional annotations. In this work, we propose a novel interpretable-by-design method, ProtoVerse, to find relevant sub-parts of vertebral fractures (prototypes) that reliably explain the model's decision in a human-understandable way. Specifically, we introduce a novel diversity-promoting loss to mitigate prototype repetitions in small datasets with intricate semantics. We have experimented with the VerSe'19 dataset and outperformed the existing prototype-based method. Further, our model provides superior interpretability against the post-hoc method. Importantly, expert radiologists validated the visual interpretability of our results, showing clinical applicability.
	
\end{abstract}



\section{Introduction}
\label{sec:introduction}

Osteoporosis is one of the most prevalent diseases among elderly people, which is clinically manifested as bone fractures localised typically in the spine, hip, distal forearm, and proximal humerus. These fractures can lead to severe pain, kyphosis, disability, as well as approx. 12-fold increased risk for further fractures and 8-fold increased mortality \citep{cauley2000risk}. If such fractures are detected timely, they can be effectively treated to reduce further fracture occurrences and morbidity.
Unfortunately, most such cases go undetected until the occurrence of the first acute symptoms.  Vertebral Compression Fractures (VCFs) are the most prevalent among osteoporotic fractures occurring in the vertebrae. VCFs inflict $30-50 \%$ population above the age of 50 \citep{ballane2017worldwide} and have been reported to be undiagnosed in $84 \%$ cases via Computed Tomography (CT) \citep{carberry2013unreported}. 
In clinical practice, radiologists primarily rely on inspecting the sagittal plane of CT scans for detecting VCFs. Moreover, radiologists grade the detected VCF using \textit{Genant scale} \citep{genant1993vertebral}, which measures the reduction in vertebral height from CT images. The scale has four grades: G0 (Healthy), G1 (Mild/ $< 25 \%$ height reduction), G2 (Medium/ $25-40 \%$ reduction), and G3 (Severe/$> 40 \%$ reduction). Note that, fracture grading is a more complex task than fracture detection since it is a fine-grained classification and relies on radiologists' skill and experience. Alternatively, Deep Learning (DL) has the potential to be used as an assistive tool for overloaded radiologists to accelerate VCF detection and VCF grading. However, the conventional DL-based method offers slim to none interpretability, making its deployment unreliable in safety-critical systems such as clinical practice. In order to effectively reduce the burden of radiologists and facilitate faster diagnosis, human-interpretable DL methods are the need of the hour. In this paper, we aim to develop a novel interpretable-by-design DL model for explainable VCF grading.
\begin{figure}[t]
    \centering
     \includegraphics[width=0.95\textwidth]{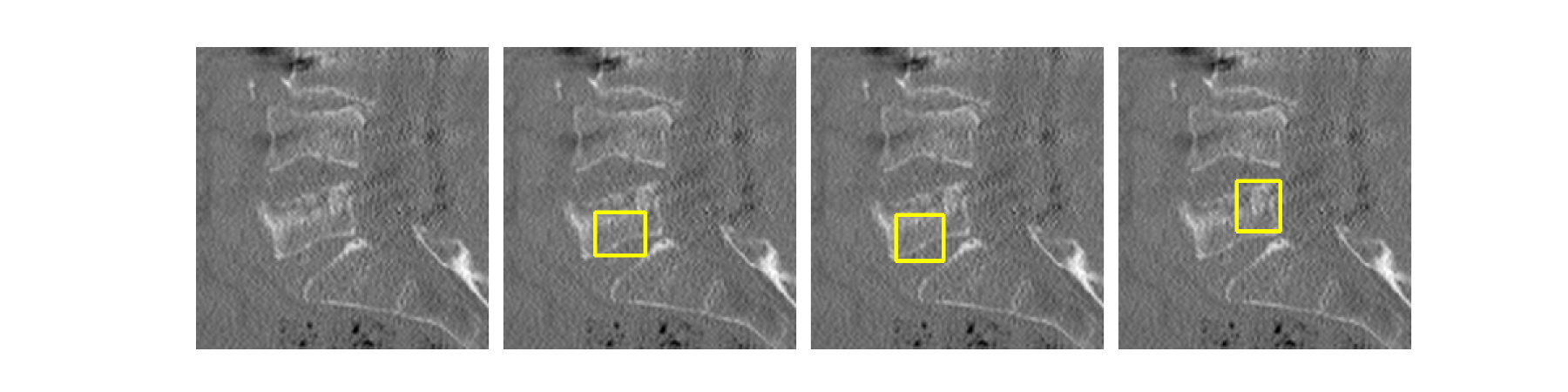}
   \caption{Prototype-activated regions (yellow boxes) on the fractured vertebra of a typical test sample provides human-interpretable reasoning for its fracture grade G3.
   }
   \label{fig:prototypes}
\end{figure}
Recently eXplainable AI (XAI) methods have been developed to 
estimate decision-making factors of Deep Neural Networks (DNNs) into human-interpretable formats from the learnt representations or activations. However, what makes an explanation sufficiently human-interpretable is largely subjective of the XAI method, use-case, and the dataset in question \citep{sinhamahapatra2023towards}. While some initial approaches rely on post-hoc explanations, recent methods are designed to be inherently interpretable.
For critical healthcare applications, it is important that we generate explanations that are aligned with how doctors would interpret them. A growing direction in current XAI approaches is to learn representations that can be tied directly to higher level \textit{human-understandable concepts}. An example from natural images could be predicting an image of a \textit{red-billed hornbill}, which depends on the presence of concepts like \textit{red bill}. 
However, an explicit identification of such concepts requires costly expert annotations for each sample and often can be non-exhaustive. Instead, one could learn such concepts in an unsupervised manner as \textit{prototype} parts, which are semantically recurring features for a given class in a dataset. These prototypes are representative parts sufficient to identify the given class, e.g., if prototype parts of \textit{beaks, wings, tails} are learnt, they could together reliably predict a \textit{bird}. Similarly, in VCF fracture grading (c.f. Fig. \ref{fig:prototypes}), detecting G3 grade fracture relies on severe deformities in the vertebral body. While learning such prototype representations in an unsupervised manner is a challenge by itself, visualising such implicitly learnt representations in a human-understandable way is also difficult. Recently \textit{Prototypical Parts Network} (ProtoPNet) \citep{chenThisLooksThat2019a}, has been introduced that enables DNNs to learn internal representations as a set of interpretable prototypes consisting of parts of the closest training image. During inference, parts of test samples could be visually reasoned to be similar (`this-looks-like-that') to some of its closest prototypes. 
In this work, we leverage prototype-based learning for human-interpretable VCF grading. By learning visual prototypes that sufficiently represent different fracture grades, one can explain the inferred VCF grading of a test sample using its similarity with the prototypes from that fracture grade (c.f. Fig. \ref{fig:prototypes}). Such an explainable design would bring trustworthiness to the deployed system and reduce the workload of clinicians and diagnosis time for patients. However, existing prototype-based models struggle to capture necessary visual and semantic variations within the prototypes of the same class. Repetition in prototypes can result in inadequate human-interpretable visual explanations during inference. Crucially, these methods were designed to target natural images where the semanticity is intuitive and abundant. In comparison, semanticity in medical images is intricate and involves a complex, expert understanding of the normality of anatomical patterns, including edges, shapes, volume, and local contrast. Hence, these models are more prone to visual feature collapse in the prototype layer for the medical dataset, which calls for appropriate customisation. This is even exaggerated in the case of small dataset size.
We mitigate this by a novel diversity-promoting loss, which forces the network to focus on a diverse set of vertebral features, e.g., various fracture regions. Further, we employ imbalance-aware classification loss to alleviate dataset imbalance. In the end, our proposed method is suitably applied for VCF grading in an imbalanced and small dataset.

\vspace{0.5em}
\noindent \textbf{Contributions:}  In summary, our contributions are:
\vspace{-0.5em}
\begin{itemize}
    \item We propose a novel interpretable-by-design method \textit{ProtoVerse}, for VCF grading based on prototype parts learning. To the best of our knowledge, we are the first to leverage interpretability based on prototypes into clinical VCF datasets.
    \vspace{-0.5em}
    \item We propose a novel objective function to enhance the visual and semantic diversity of the learnt prototype for each fracture / healthy class. Additionally, we mitigate huge class imbalances with the help of differential weighting in the classification loss.
    \vspace{-0.5em}
    \item Our method substantially outperforms ProtoPNet, a state-of-the-art prototype-based method. Moreover, our method precisely localises clinically relevant parts in the vertebral body, while a prominent post-hoc method, GradCAM, fails to achieve this.
\vspace{-0.5em}
    \item Finally, we demonstrate the clinical applicability of our method by validating its human-interpretability with the help of expert radiologists.
    
\end{itemize}

\section{Related Work}
\label{sec:related_work}

\noindent\textbf{Human-interpretable DL Models:} 
A collection of XAI works for visual tasks can be found in the surveys 
\citep{nguyenUnderstandingNeuralNetworks2019, samekExplainableArtificialIntelligence2019} including the medical application \citep{tjoaSurveyExplainableArtificial2020}. The explanations can be categorised based on several criteria, such as (a) local vs. global, (b) post-hoc vs. interpretable-by-design (IBD), and (c) explicitly specified vs. implicitly derived. Depending upon the user requirement, a particular XAI approach can be used for generating explanations. \textit{Local} explanations deal with single sample by looking at its specific parts with highest attribution for a particular decision in DNNs, such as saliency-based class activation map (CAM) \citep{zhouLearningDeepFeatures2016}, GradCAM \citep{selvaraju2017grad} and its variants like GradCAM++ \citep{Chattopadhyay2017GradCAMGG} and XGradCAM \citep{Fu2020AxiombasedGT}, gradient-based local sensitivity, \citep{sundararajanAxiomaticAttributionDeep2017} and relevance back-propagation \citep{bachPixelWiseExplanationsNonLinear2015}. 
Previous studies have pointed out that saliency-based methods for post-hoc explainability are limited under constant intensity shift \citep{kindermans2019reliability}, model weight randomization \citep{arun2021assessing}, relatively smaller regions of interest \citep{saporta2022benchmarking}, confounders \citep{sun2023right}, and often disagree with medically correct explanation \citep{rafferty2022explainable}.
In contrast, \textit{global} explanations consider the whole dataset and generate explanations based on recurring features in multiple samples. 
Global explanations derived from such learnt internal representations are more useful and often necessary for tasks like anomaly detection.
Most local methods generate \textit{post-hoc} explanations from already trained models. Since these explanations are generated a posteriori, they are sub-optimal to make use of high-level semantic recurrence in the internal representation. Therefore, in order to make the internal representations human-interpretable,  we need to consciously design global methods that are inherently \textit{IBD} \citep{rudinStopExplainingBlack2019}. Here, one recent approach maps human-understandable concepts to internal representations as an interpretable bottleneck layer \citep{schwalbeConceptEmbeddingAnalysis2022}. These concepts can be either \textit{explicitly specified} per sample as attributes in natural language by domain experts as in concept learning models \citep{kohConceptBottleneckModels2020a, kazhdanNowYouSee, fongNet2VecQuantifyingExplaining2018, fangConceptbasedExplanationFinegrained2020, caoConceptLearnersFewShot2021}
or \textit{implicitly derived} as in prototype-based learning models \citep{chenThisLooksThat2019a, nautaNeuralPrototypeTrees2021, gautamThisLooksMore2021, liDeepLearningCaseBased2018, donnellyDeformableProtoPNetInterpretable2022a, rymarczykInterpretableImageClassification2022}. The latter does not depend on concept-specific annotations but rather simply learns to associate latent representations to parts of the closest training images. These parts are termed as \textit{prototypes} \citep{chenThisLooksThat2019a}, which serve as semantically relevant visual explanations. 
Subsequent works have pointed out multiple limitations, such as the semantic gap between latent and input space \citep{hoffmann2021looks},  and the distance of prototypes from the decision boundary \citep{wang2023learning}. Since then, many works have extended and improved prototypical learning by imposing consistency and stability criteria \citep{huang2023evaluation}, using support prototypes \citep{wang2023learning}, mining negative prototypes \citep{singh2021these}, and receptive field-based localization \citep{carmichael2024pixel}.

\vspace{0.5em}
\noindent \textbf{Vertebral Fracture Detection:} 
Numerous recent works have studied vertebrae fracture detection on both the patient and vertebrae levels; however, only a few have attempted to make the network decision IBD. The CNN-based networks take center stage in the architectural choice, which is used in both 2D slices and 3D volumes of CT images. Among 2D-based works, a custom contrastive loss based on grading hierarchy to process the reformatted mid-sagittal 2D slices was proposed in \citep{husseiniGradingLossFracture2020}. In \citep{pisov2020keypoints}, a key points based method was proposed to localize vertebrae and subsequently estimate fracture severity. An LSTM-based approach was used in  \citep{tomita2018deep} and \citep{bar2017compression} to leverage inter-slice and intra-slice feature dependencies, respectively.
While 2D methods can efficiently capture discriminative features of vertebrae, 3D methods explicitly target 3D context at the cost of increased computational cost. A 3D dual-pathway architecture for dense voxel-wise fracture detection was proposed in \citep{nicolaes2020detection}. A sequence-to-sequence CNN for inter-patch feature accumulation was used in \citep{chettrit20203d}. A patient-level feature aggregation from 3D patches was proposed in \citep{yilmaz2021automated}. A two-stream architecture to explore discriminative features for fine-grained classification was used in \citep{feng2021two}.

Recently, \citep{engstler2022interpretable} proposed to leverage an activation map to find correlated image regions to a fracture grade by expert validation. 
However, their method offers post-hoc interpretability and heavily depends on the post-processing heuristics.
In stark contrast, we aim to obtain a global IBD model 
which enforces the network to learn the most semantically recurring features across the dataset. \citep{zakharov2023interpretable} proposes a fracture grading pipeline, which provides interpretability based on the measured heights between landmarks of vertebrae. While this approach closely follows the Genant definition, it requires additional annotation of six key points per vertebrae, making the annotation process complex and more time-consuming.

Few recent approaches applied prototype-based learning methods in medical use-cases, such as Alzheimer’s disease \citep{mohammadjafariUsingProtoPNetInterpretable2021}, brain tumor classification \citep{weiMProtoNetCaseBasedInterpretable2023}, and breast masses classification \citep{carloni2022applicability}. However, their goal is limited to binary classification, such as anomaly detection (healthy vs. unhealthy).
In comparison, our method brings interpretability to fine-grained multi-class medical classification tasks such as VCF grading, where small sample sizes and severe class imbalances pose additional challenges. Previously, \citep{kimXProtoNetDiagnosisChest2021} has shown multi-class use cases on chest X-rays; however, their approach is tailored for disease with dynamic areas of focus.

\section{Dataset}
\label{sec:dataset}

In this work, we use the VerSe'19 dataset from the VerSe 2019 challenge for VCF grading. VerSe \citep{sekuboyina2021verse} is a large-scale, multi-detector, multi-site CT spine dataset comprising 374 scans from 355 patients. To minimize existing inter-rater variability of publically available annotation \citep{loffler2020vertebral}, we updated annotations in-house by a group of radiologists with 15 years of median experience. 
For VCF, in this work, we consider healthy, G2 and G3. Following \citep{husseiniGradingLossFracture2020, keicherSemanticLatentSpace2023, engstlerInterpretableVertebralFracture2022}, we do not consider G1 since it is often clinically challenging to differentiate G1 from healthy class and is prone to significant inter-rater variability. We have altogether 1444 vertebrae annotation; out of 1308 are healthy, 76 are G2, and 52 are G3. We have 1086 training samples and 358 test samples. We have 326 healthy, 18 G2, and 14 G3 test samples. Clearly, our main challenge is to devise a useful learning strategy for a successful interpretable DL method to mitigate high class imbalance and small sample size.

\section{Methodology: ProtoVerse}
\label{sec:method}

In this section, we propose a novel interpretable method called ProtoVerse for vertebral fracture grading. At the core of this lies an interpretable architecture motivated by ProtoPNet \citep{chenThisLooksThat2019a}, which is an image classifier network that learns inherent representations for relevant sub-parts of an image. Once these representations are learnt, they serve as visual explanations for the model output. To obtain such visual explanation, sub-parts or patches of training images closest to the learnt representations are stored in a set of visual patches called \textit{prototypes} to enhance human-interpretation of the inner workings of DNN models.

\begin{figure}
    \centering
    \includegraphics[width=0.9\textwidth]{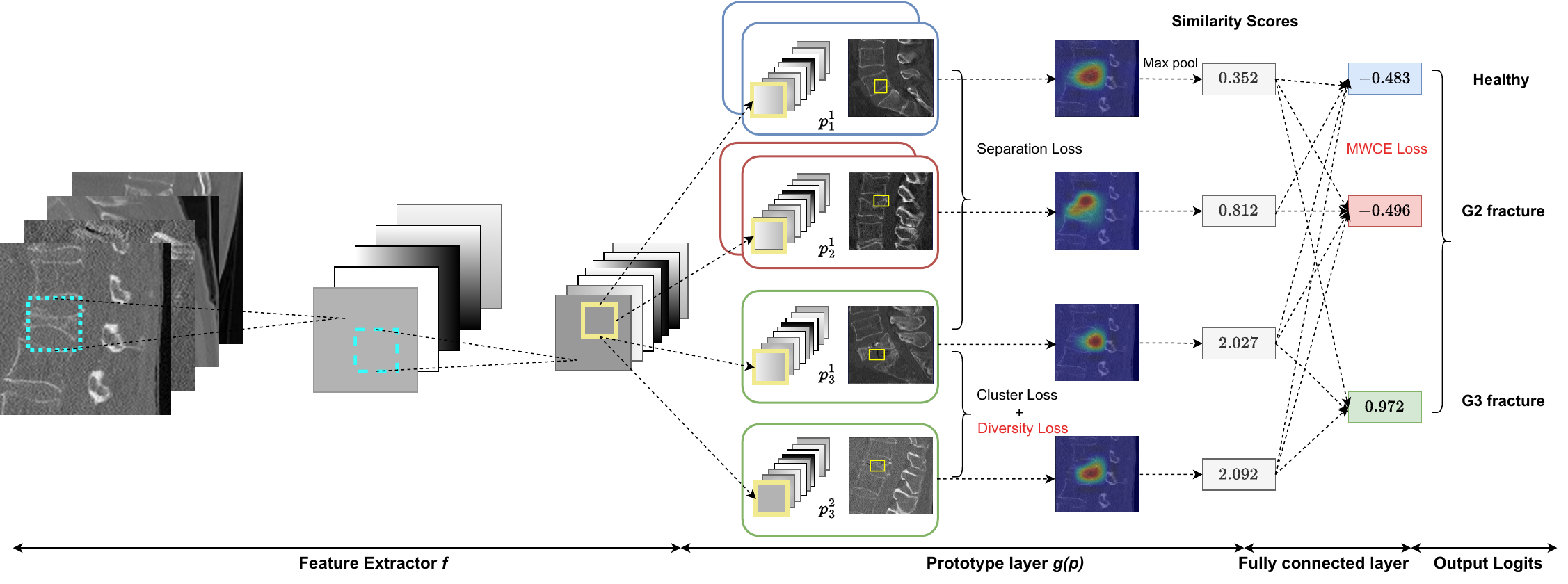}
    \caption{ProtoVerse architecture for learning prototypes for VCF grading. Prototypes from each class are shown as : healthy (blue), G2 (red), and G3 (green), which learns representative image parts for each class through separation and clustering loss. For example, healthy prototypes emphasise straight vertebral edges, while G2 and G3 prototypes capture the different degrees of deformities in vertebrae. Notably, our novel Diversity loss ensures the capture of visual variations within a class such as  $\boldsymbol{p}_{3}^1$ and $\boldsymbol{p}_{3}^2$ highlighting different fracture regions in G3.
    Given a G3 input test sample, the prototype patch $\boldsymbol{p}_{3}^1$ and $\boldsymbol{p}_{3}^2$ belonging to G3 shows the strongest presence (similarity score $2.027$ and $2.092$) in various fracture regions for vertebrae of interest. While the prototype parts $\boldsymbol{p}_{2}^1, \boldsymbol{p}_{1}^1$ belonging to G2 and healthy classes have lesser similarity score. Final classification logits are trained with our MWCE loss to mitigate class imbalance.} 
    
  \label{fig:protoverse_arch}
\end{figure}

Figure \ref{fig:protoverse_arch} shows the overview of the \textit{ProtoVerse} for learning visual prototypes for vertebral fractures. The model consists of a convolutional feature extractor backbone $f$ followed by two additional $ 1 \times 1$ layers, a prototype layer $g_p$, and a fully connected layer $h$ with weights $w_h$. The feature extractor is usually taken from standard Convolutional Neural Networks (CNNs), e.g., ResNets, DenseNets, and VGG. These CNNs are initialised using ImageNet pre-trained weights. During training, firstly, we learn the prototype vectors in the latent space via stochastic gradient descent over the objective function until the last layer. Next, the last layer for classification is updated by a convex optimisation. Finally, the learnt prototype vectors are visualised through retrieved \textit{prototype} patches/sub-parts of training images. In the following, we expand upon each of the working stages of ProtoVerse.

\subsection{Learning Prototype Vectors}
\label{sec:method_learning}

Let us consider an input image $\boldsymbol{x}$ with a  fracture grade $\in \{G0, G2, G3\}$. The feature extractor generates a convolutional output $\boldsymbol{z} = f(\boldsymbol{x})$ with a shape of $\boldsymbol{z} \in \mathbb{R}^{H \times W \times D}$, where $D$ is the feature size in the last convolutional layer. Given the feature extractor is ImageNet pre-trained, the input image is  resized to $224 \times 224 \times 3$. The ouput of the $f$ produces $7 \times 7$ patches, i.e. $H=7, W=7$.

Next, the output of $f$ goes through the prototype layer $g_p$ consisting of learnable prototype vectors $\mathbb{P}=\{\boldsymbol{p}_j^k\}_{j=1:c,k=1:m}$, where $m$ is the number of prototypes per class and $c$ is the total number of classes. Thus, the total number of prototypes $n=m\times c$. Each prototype $\boldsymbol{p}_j^k \in \mathbb{R}^{h \times w \times D}$ has the same feature size as the last convolutional layer. Following ProtoPNet \citep{chenThisLooksThat2019a}, we have taken prototypes with $h = w = 1$. Thus, each prototype corresponds to a prototypical activation pattern $1 \times 1 \times D$ patch in the $ 7 \times 7 \times D$ last-layer feature. Since the $7 \times 7 \times D$  feature vector represents the entire input image, a prototype vector $\boldsymbol{p}_j^k$ corresponds to a proportionate field of view of a single feature in a patch, which is considered as prototype patch/sub-part of the input image. Thus, $n$ number of prototype vectors corresponds to the $n$ most salient image parts characterising the given dataset.

Next, we calculate the $\ell_2$ distances between each of the prototype $\boldsymbol{p}_j^k$ and all the similar-sized patches of the feature output layer $\boldsymbol{z}$. The smaller the distance, the higher the similarity of the prototype vector to the feature patch for the given input image. Thus, the point-wise inversion of the $\ell_2$ distances provides $7 \times 7$ activation map referred to as \textit{similarity heatmap}.  To reduce the similarity heatmap into a single \textit{similarity score}, global max pooling is performed. This helps in interpreting how strongly a particular prototype vector matches with a specific part of the input image. Thus, the prototypes for a particular class $j$ are responsible for capturing the most relevant parts for images of class $j$. Lastly, in the fully connected layer $h$, the $m$ similarity scores from each class are multiplied with the weight matrix $w_h$, which constitutes the \textit{class connection score}. For an input $x_i$, the output logits are obtained as $\hat{y}_i = h \circ g_p \circ f(x_i)$. The resultant outputs are normalised using softmax to produce the probabilities for images belonging to various classes.

To learn meaningful prototype vectors in the latent space, prototypes of each class should be clustered together and well-separated from the prototypes of different classes. ProtoPNet \citep{chenThisLooksThat2019a} tried to achieve this goal by introducing intra-class clustering and inter-class separation loss, in addition to classification loss.
For a given training dataset, $\mathbb{D}_{\mbox{train}} = \{\boldsymbol{x}_i, y_i\}^{d}_{i=1}$ and a set of prototype vectors for $j^{\mbox{th}}$ class $\mathbb{P}_{y_i} = \{\boldsymbol{p}_{y_i}^k\}^{m}_{k=1}$, the cluster loss ($\mathcal{L}_{\mbox{clst}}$) and separation loss ($\mathcal{L}_{\mbox{sep}}$) is as follows: 
\begin{eqnarray}
    \mathcal{L}_{\mbox{clst}} = \frac{1}{d}\sum_{i=1}^d\min_{j:\boldsymbol{p}_j^k \in \mathbb{P}_{y_i}} \min_{\boldsymbol{z} \in \chi(\boldsymbol{z}_i)} \| \boldsymbol{z}-\boldsymbol{p}_j^k\|_2^2 \label{eq:loss_clst}; \quad
    \mathcal{L}_{\mbox{sep}} =  \frac{1}{d}\sum_{i=1}^d\min_{j:\boldsymbol{p}_j^k \notin \mathbb{P}_{y_i}} \min_{\boldsymbol{z} \in \chi(\boldsymbol{z}_i)} \| \boldsymbol{z}-\boldsymbol{p}_j^k\|_2^2 \label{eq:loss_sep}
\end{eqnarray}

where $\chi(\boldsymbol{z}_i)$ denotes all the patches from feature $\boldsymbol{z}_i$.Separation loss pushes apart prototypes of different classes in the prototype layer. On the contrary, cluster loss tries to increase semantically similarity among prototype patches of its own class based on $\ell_2$ distances. At the same time, the weights of the fully connected layer are initialised following \citep{chenThisLooksThat2019a}. Specifically for a given class, we initialise class connection for the prototype of the same class as $1.0$ and $-0.5$ for prototypes of a different class.


Since VerSe'19 is a relatively small dataset, it is increasingly challenging to learn relevant prototypes for fracture classes. Particularly, cluster loss in Eq. \ref{eq:loss_clst} encourages prototypes within the same class to be close to each other. However, due to the absence of many diverse samples, this inevitably leads to prototypes that are semantically similar and repeating patches. For a more interpretable diagnosis, it is often preferred to look at different patches of the given fracture grade, for e.g., prototypes focusing on fractures on the top/ bottom edge or the frontal/ posterior vertical edges of the vertebrae. 
To this end, we propose a novel loss to promote the diversity of prototypes for a given class.
Moreover, VerSe'19 has only $5 \%$  fracture samples compared to healthy samples. With such a heavy class imbalance, the classification and Separation loss tends to be skewed towards healthy class.
Since interpretable learning demands a sufficient number of samples, we aim to keep all available healthy vertebrae and employ a class imbalance aware loss function.
Next, we describe our proposed training strategies comprising novel prototype diversity loss and imbalance-aware classification loss for better interpretability.


\subsubsection{Novel Prototype Diversity Loss}
\vspace{-0.5em}
Given very few samples from fracture classes, only a limited variation of visual parts showing predominant characteristics for the fracture classes are available to learn the prototypes. Further, semantics in medical images involve an expert understanding of anatomical patterns, including edges, shapes, and local texture. 
However, these features are repetitive across different scales and image locations. For example, vertebral process regions in the posterior part could have similar geometric and texture features in an anterior vertebral body. Under such constraints, clustering loss makes the network susceptible to prototype repetition. Therefore, an explicit constraint in the form of diversity loss is helpful to enforce the network to cover fine-grained variations of edges, shapes, and local textures within a certain class.
Thus, we propose a novel loss component called \textit{Diversity Loss} ($\mathcal{L}_{\mbox{div}}$), which minimizes the absolute cosine distance between prototype vectors of the same class. In action, Diversity loss prevents the collapsing of multiple prototypes of the same class into identical ones. This is achieved by balancing the weight between Cluster and  Diversity loss. As a result, the network is encouraged to find prototypes as the closest match to semantically diverse parts of the vertebrae. Experimentally, we find Diversity loss significantly mitigates the problem of redundancy \citep{sinhamahapatra2023towards} and improves the semantic and visual diversity of the prototypes. For example, in Figure \ref{fig:protoverse_arch} the prototypes $\boldsymbol{p}_{3}^1, \boldsymbol{p}_{3}^2$ from Grade $3$ fracture focus on two distinctly different parts. 
It is given by the squared mean of the dot product of the prototype vectors for a given class with a coefficient ($\lambda_{\mbox{div}} $) that decides the weight of this loss in the total loss.
\begin{equation}
    \mathcal{L}_{\mbox{div}} = \frac{2}{cm  (m-1)}\sum_{j=1:c}\sum_{k,l=1:m, k\neq l}({\boldsymbol{p}^k_j}^\top\times {\boldsymbol{p}^l_j})^2
\end{equation}
where m is the number of prototypes per class, and c is the total number of classes.

\subsubsection{Class imbalance aware classification loss}
\vspace{-0.5em}
With imbalanced dataset, the class-wise accuracy is heavily affected by model's bias towards dominant class despite achieving high sample-wise accuracy.
This leads very few samples from fractured classes to make a positive connection with the fully connected layer and in turn the prototypes learnt for these classes are not sufficiently representative for the given class. In order to provide more opportunity for learning from the samples of fractured class, we adopt differential weighting in the CE loss. However, choosing correct weighting strategy is crucial and varies with dataset size and complexity. Besides resampling strategies for tackling imbalance, we have experimented with other weighting strategies such as Inverse Number of samples (INS), Inverse of Square Root Number of Samples (ISNS). We found median weighting to be the most beneficial in our case. We have employed median-weighted cross-entropy (MWCE) in the classification loss and this has significantly improved the class-wise accuracy. Given $d_j$ represent the number of training samples from $j$-th class, the weight for the $j$-th is calculated as follows:
\begin{equation}
    \mathcal{L}_{\mbox{MWCE}} (\hat{y}_i , y_i) = -  \sum_{j=1}^{c} w_j \cdot y_i^j \log(\hat{y}_i^j); ~\mbox{where}~w_j = \frac{median(\{d_l\}_{l=1:c})}{\sum_{l=1:c ; l \neq j} d_l} \label{eq:median_weight}
\end{equation} 
Finally, we obtain the total loss function for our ProtoVerse model as follows:
\begin{dmath}
    \mathcal{L}_{\mbox{Total}} =  \frac{1}{d} \sum_{i=1}^{d} \left(\mathcal{L}_{\mbox{MWCE}} + \lambda_{\mbox{clst}} \mathcal{L}_{\mbox{clst}} + \lambda_{\mbox{sep}} \mathcal{L}_{\mbox{sep}} + \lambda_{\mbox{div}} \mathcal{L}_{\mbox{div}} \right) \label{eq:loss_ppnet}
\end{dmath}


\subsection{Visualising Prototype Vectors}
\label{sec:method_visualising}

In total, we have $n$ number of $D$-dimensional prototype vectors in latent space which is commonly visualised as clusters via various visualisation methods like TSNE, UMAP etc. However, to make our models interpretable, we need to visualise which part of input images are reflected into those latent prototype vectors. 
To address this, we follow \citep{chenThisLooksThat2019a} and extract the exact patch or image sub-part that activates the most in similarity score heatmap for each prototype vector. Such obtained image part level visualisations provide detailed  interpretability in a human-understandable manner. For each prototype vector, the visual prototype patches are extracted based on the closest training image patch from each class based on the minimum $\ell_2$ distances as given in Section \ref{sec:method_learning}. The above operation of finding the closest patch for all given training image patches for the class is given as below: 
\begin{equation}
    \boldsymbol{p}_j^k  \leftarrow  \arg   \underset{\boldsymbol{z} \in \chi(\boldsymbol{z}_i)}{min} \| \boldsymbol{z} - \boldsymbol{p}_j^k \|_2; \forall i \mbox{ s.t. } y_i = j
\end{equation}

However, prototype patches saved are not exactly the upscaled size corresponding to $1\times 1\times D$ feature patch. Instead the entire similarity heatmap for the chosen training image is upscaled to the input image dimensions. Finally, those image regions are selected as prototype patch which enclose atleast $98 \%$  in the similarity heatmap. Thus, the prototypes generated during training, act as visually interpretable parts crucial for a given class. During inference, similarity heatmap highlight the parts of a test sample that are close to the learnt prototype for a given class.

\section{Experimental Results and Discussion}
\label{sec:exp}
\vspace{-0.5em}

In this section, we demonstrate how prototype-based learning methods can be applied for enhanced interpretability to the vetebrae fracture grading usecase. We first provide details about dataset preparation experimental setup and then show our quantitative and qualitative results. Importantly, we provide insights from clinicians by conducting experiments to find the usefulness of such a method for the interpretability of fracture grades.
\vspace{0.5em}

\noindent\textbf{Data Preparation: } We adopt similar data preparation steps as \citep{husseiniGradingLossFracture2020}. Since radiologists prefer sagittal views, we gather information from 3D CT volume into reformatted 2D sagittal slices. Specifically, we construct a spline along the vertebral centroids and extract 2D reformation of the vertebrae for the sagittal plane along which the spline goes through. As a result, we have a 2D reformatted slice along the mid-vertebral plane. 
Finally, we crop a 112x112 size region of interest comprising valuable context of at least one vertebra above and below, keeping the vertebrae of interest at the center, and subsequently resize the image to 224x224. We have not noticed any significant difference in accuracy drop in training from scratch for different image resolutions [112x112, 224x224]. However, for improved performance, we have always used an imagined-trained backbone in our experiments, and hence we have resized it to 224x224.
Moreover, we provide an additional channel with a Gaussian around the centroid of the vertebra of interest.

\vspace{0.5em}
\noindent\textbf{Experimental Setup: } We have used similar settings as in \citep{chenThisLooksThat2019a} for model training and prototype visualisation. In our experiments, we select a classifier without a prototype layer as a baseline for direct comparison between post-hoc interpretability of the baseline with our proposed interpretable-by-design method. 
Since previous methods \citep{husseiniGradingLossFracture2020, keicherSemanticLatentSpace2023} used different fracture annotations than ours, their results cannot be used for direct comparison with our model. We believe our choice of baselines with and without IBD serve as sufficient points of comparison in our experiments.
Further, we show ablation of our proposed diversity loss. We have extensively evaluated with various pre-trained backbones such as VGG \citep{simonyan2014very}, ResNet \citep{he2016deep}, and DenseNet \citep{huang2017densely} and found that VGG-11 results in the best performance.
We have also conducted an ablation study on the number of prototypes per class and found the performance decreasing with an increasing number of prototypes (Sec \ref{sec:ablation}). 

To quantitatively evaluate the performance of imbalanced dataset like VerSe'19, we have focused on \textit{Class Average Accuracy, Class Average F1} scores as evaluation metrics, such that performance on individual classes are highlighted as compared to \textit{Sample Accuracy} based on total number of correct samples.

\begin{table}[th]
    \centering
    \begin{tabularx}{\linewidth}{lllll}
        \toprule
        \textbf{Models}   & \textbf{IBD} & \textbf{Class-avg} & \textbf{Class-avg} & \textbf{Sample} \\
                           & & \textbf{Accuracy} & \textbf{F1} & \textbf{Accuracy} \\
        \midrule
        Baseline          & \xmark            & 72.46                           & 60.82                     & 86.03                      \\
        ProtoPNet \citep{chenThisLooksThat2019a} & \cmark            & 64.81                           & 51.59                     & 83.24                     \\
        ProtoVerse (ours) & \cmark            & \textbf{76.28}                           & \textbf{67.97}                    & \textbf{90.22}                    \\ \bottomrule
    \end{tabularx}
    \caption{Comparison of our ProtoVerse model with ProtoPNet and a baseline classifier on the test set. All models have been trained on Verse'19 dataset using MWCE for class imbalance. ProtoVerse, thanks to novel Diversity loss, significantly outperforms ProtoPNet and non-IBD baseline.}
    \label{tab:accuracy-table}
\end{table}
\subsection{Classification Results}
\label{sec:results_classification}

We use 10\% of the training data as a validation set for model selection and a fixed held-out test set. Our main results on the test set are presented in Table \ref{tab:accuracy-table}, and clinical evaluation is performed on this test set.
In Table \ref{tab:accuracy-table}, we compare our ProtoVerse classification results with ProtoPNet and the post-hoc baseline model.
For all the models, we used pre-trained VGG11 as the backbone and MWCE for classification loss since it was found to be a better fit for the imbalance in the dataset. Using the MWCE for classification, we could ensure reasonably good class-average test accuracies even for such a highly imbalanced dataset. The class-average accuracy for each class for each of the models are given as -
Baseline: $[88.04, 72.22, 57.14] \%$, ProtoPNet: $[86.5, 22.22, 85.71] \%$, ProtoVerse: $[92.33, 72.22, 64.29]\%$. 
We observe that ProtoPNet does not perform as well as the baseline model in terms of the considered metrics. We attribute this to the feature collapse in the prototype layer. Importantly, we observed that the ProtoVerse model significantly outperforms both ProtoPNet and baseline across all metrics.
For our ProtoVerse model, we found  ($\lambda_{\mbox{div}} =0.3$) to be the optimum coefficient for Diversity loss. We postulate that explicitly ensuring diverse prototypes is beneficial to learn IBD models, especially in case of small sample size. The Diversity loss forces the network to focus on different parts of the vertebrae body, which is in line with clinical diagnosis rationale so that the variations in a fracture class can be adequately explained using the learnt prototypes. 

\begin{table}[th]
    \centering
    \begin{tabularx}{\linewidth}{lllll}
        \toprule
        \textbf{Models}   & \textbf{IBD} & \textbf{Class-avg} & \textbf{Class-avg} & \textbf{Sample} \\
                           & & \textbf{Accuracy} & \textbf{F1} & \textbf{Accuracy} \\
        \midrule
        Baseline          & \xmark            & 70.92   $\pm$ 3.29                        & 63.52 $\pm$ 4.68                    & 89.26   $\pm$ 3.84                   \\
        ProtoPNet \citep{chenThisLooksThat2019a} & \cmark            & 72.73 $\pm$ 4.47                           & 67.27 $\pm$ 4.06                     & 91.13 $\pm$ 3.56                     \\
        ProtoVerse (ours) & \cmark            & \textbf{76.36} $\pm$ 5.96                          & \textbf{75.97} $\pm$ 4.80                    & \textbf{93.58}  $\pm$ 1.37                  \\ \bottomrule
    \end{tabularx}
    \caption{
    5-fold cross-validation of our ProtoVerse model with ProtoPNet and a baseline classifier. Average and standard deviations over 5 runs are reported.}
    \label{tab:cross-val}
\end{table}
Additionally, we have performed a 5-fold cross-validation on the train set among non-IBD baseline, ProtoPNet, and ProtoVerse in Table \ref{tab:cross-val}. The general trend remains the same in the cross-validation as observed for the test set in Table \ref{tab:accuracy-table}. Accuracy for each class averaged across 5 splits are as follows: Baseline: $[92.75, 39.99, 80] \%$, ProtoPNet: $[93.89, 58.33, 66]\%$, ProtoVerse: $[96.732, 62.33, 70]\%$. We also performed a pairwise Wilcoxon signed-rank test to determine the statistical significance between ProtoVerse, ProtoPNet, and Baseline. For Class-avg Accuracy, we find p-values 0.094 (ProtoVerse vs. Baseline) and 0.031 (ProtoVerse vs.  ProtoPNet), respectively. For Class-avg F1, we find p-values 0.031 (ProtoVerse vs. Baseline) and 0.031 (ProtoVerse vs.  ProtoPNet), respectively. For Sample Accuracy, we find p-values 0.031 (ProtoVerse vs. Baseline) and 0.031 (ProtoVerse vs.  ProtoPNet), respectively. ProtoVerse shows statistically significant (p-value$<$0.05) results in two out of three metrics compared to baselines and all three metrics for ProtoPNet.


\begin{figure*}[!h]
    \centering
    \includegraphics[width=0.8\textwidth]{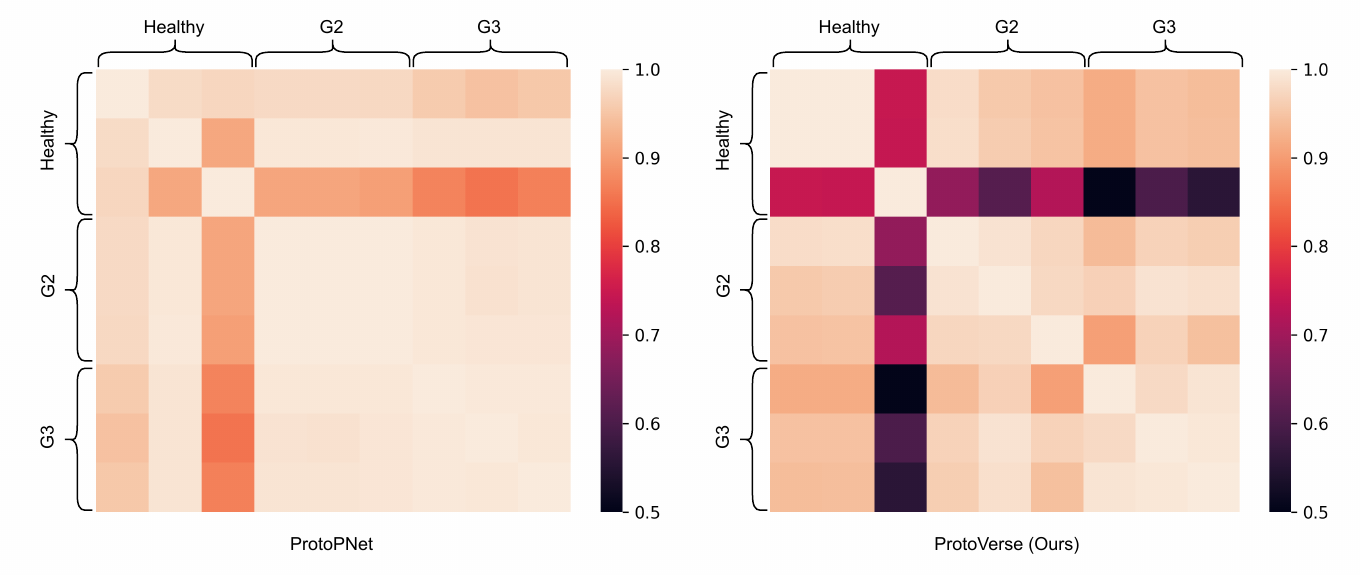}
   \caption{
   Cosine similarity between prototype vectors obtained from ProtoPNet and our ProtoVerse. Note that cosine similarity within a class is relatively high in the case of ProtoPNet, indicating difficulties in encompassing diverse prototypes. ProtoVerse achieves a diverse set of prototypes, as reflected in cosine similarity scores. Note that the third healthy prototype in both cases is looking into the trachea, which results in totally different features than vertebrae.
   }
  \label{fig:cos_sim}
\end{figure*}

\begin{figure*}[!h]
    \centering
     \begin{subfigure}{0.95\textwidth}
        \centering
        \includegraphics[width=\textwidth]{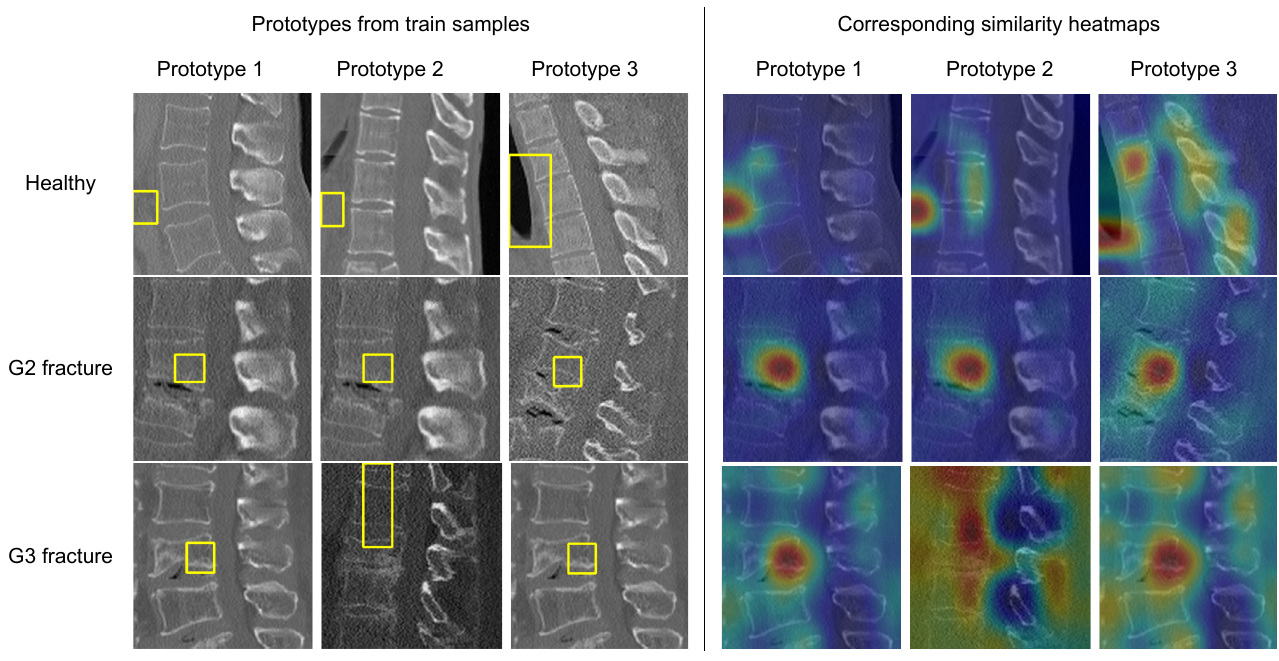}
        \caption{ProtoPNet prototypes}
    \label{fig:protopnet_prototypes}
    \end{subfigure}
    
    \vfill
    \begin{subfigure}{0.95\textwidth}
        \centering
        \includegraphics[width=\textwidth]{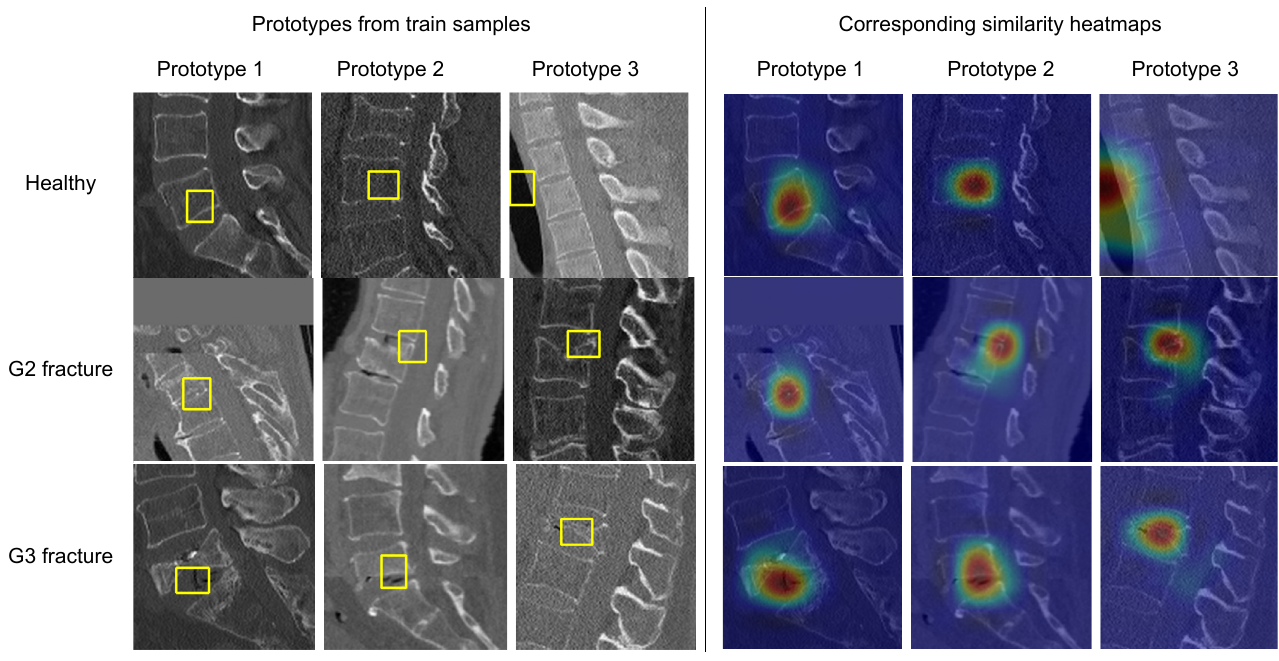}
        \caption{ProtoVerse prototypes}
    \label{fig:protoverse_prototypes}
    \end{subfigure}
      \vspace{-0.3em}
   \caption{Qualitative comparison of prototype learnt from ProtoPNet and ProtoVerse model. Note that ProtoPNet produces repetitive prototypes whereas ProtoVerse captures diverse prototypes.}
  \label{fig:prototype_comparison}
\end{figure*}

\subsection{Prototype Quality of ProtoPNet vs ProtoVerse}
We present prototypes from ProtoPNet and ProtoVerse models in Fig. \ref{fig:prototype_comparison} and cosine similarity among the prototype vectors in Fig. \ref{fig:cos_sim}.
We see that ProtoPNet produces repetitive prototypes for G2 and G3 and focuses outside the vertebral body for the healthy class. In contrast, ProtoVerse produces diverse prototypes for all three classes. Note that since most of the thoracic vertebras do not have any fracture, both models see the nearby trachea region of those cases (3rd prototype of healthy class) as an attribute of healthy class.

\subsection{Visual Interpretability Analysis}

\begin{figure*}[!h]
    \centering
     \begin{subfigure}{0.95\textwidth}
        \centering
        \includegraphics[width=\textwidth]{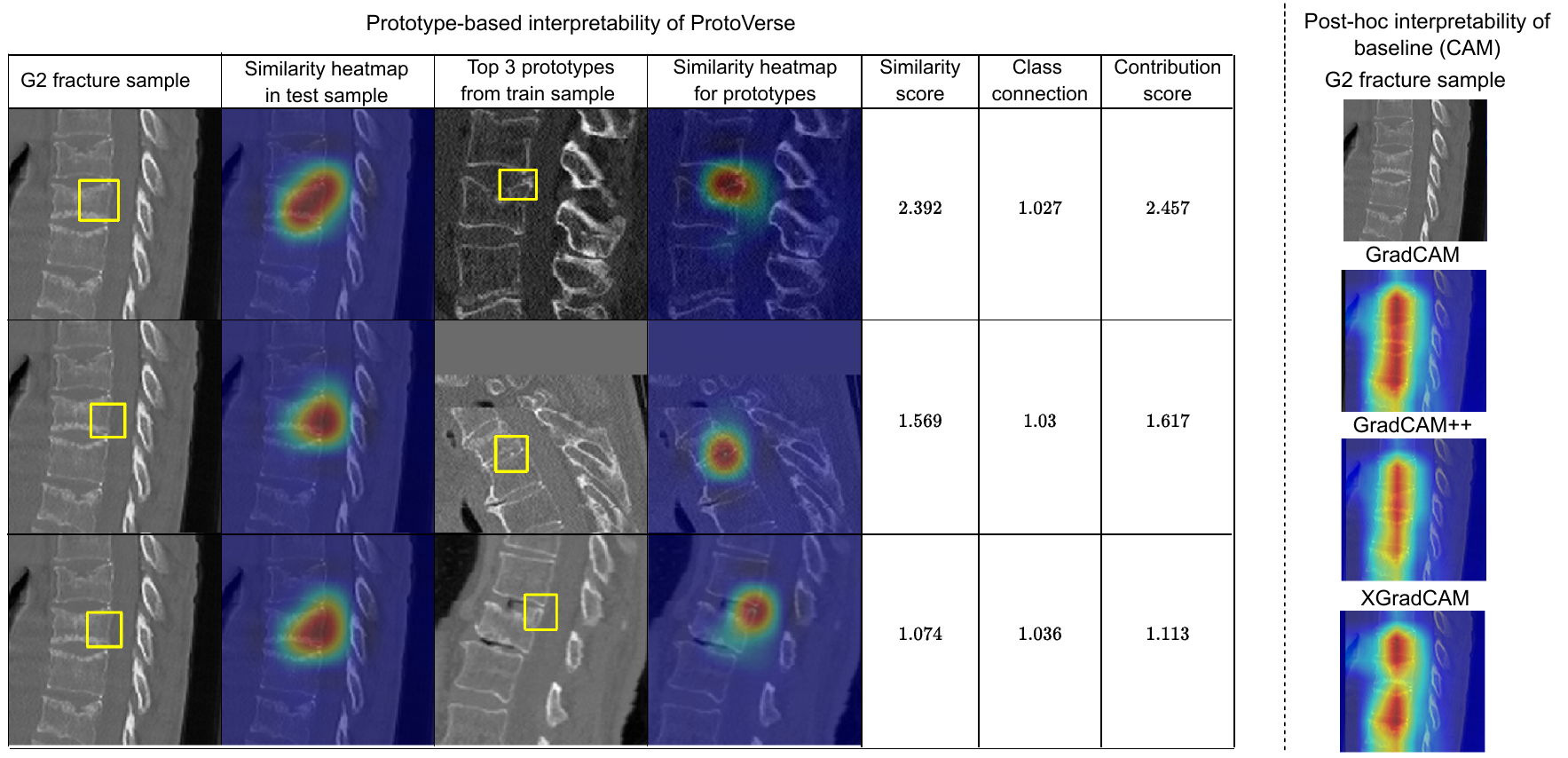}
    \end{subfigure}
    
    \vfill
    \begin{subfigure}{0.95\textwidth}
        \centering
        \includegraphics[width=\textwidth]{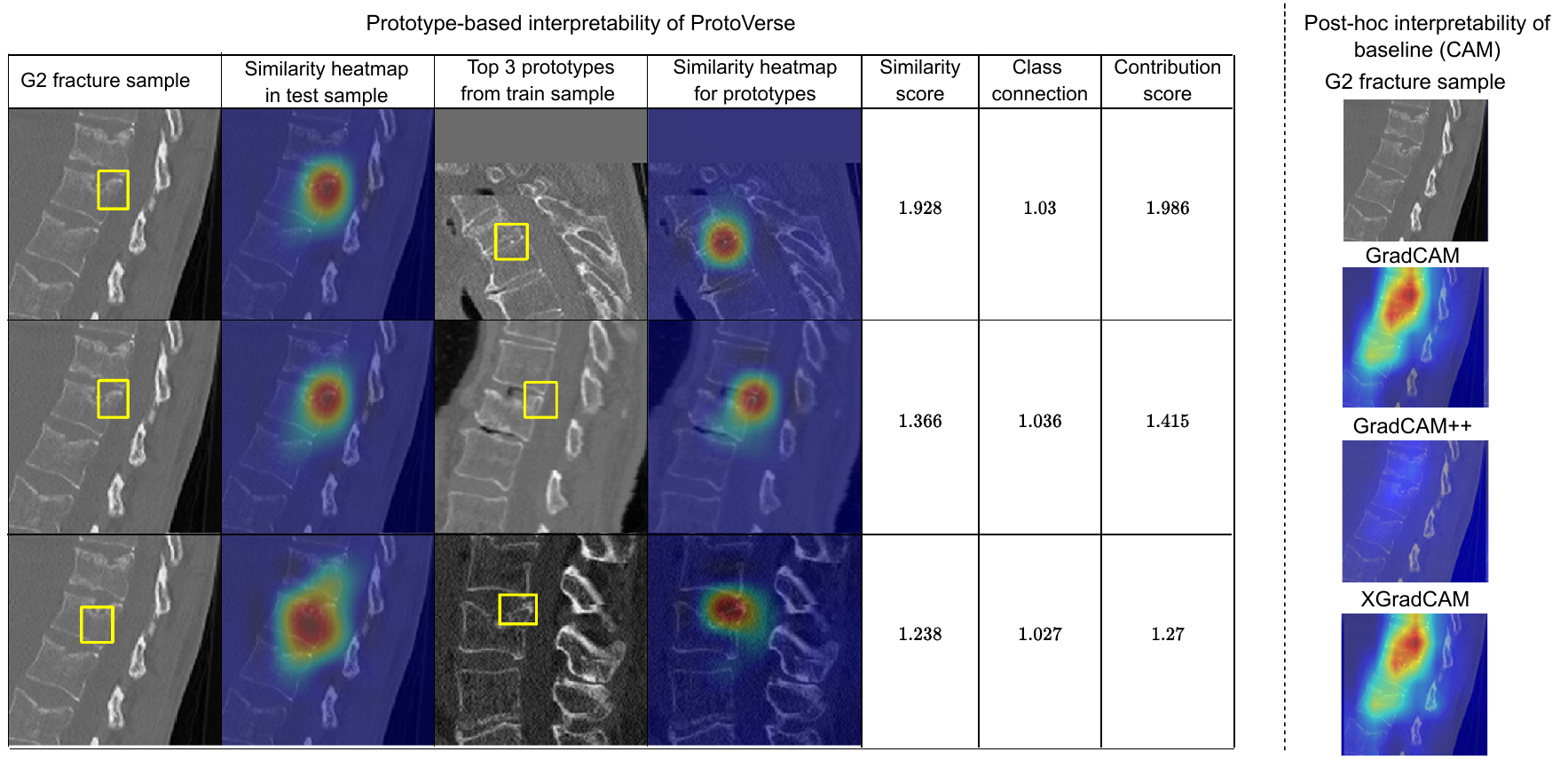}
    \end{subfigure}
      \vspace{-0.3em}
   \caption{Test-time interpretability of ProtoVerse (left) in comparison to post-hoc baselines (right) for two typical G2 fracture samples. 
   We show the top $3$ closest prototypes (col. $3$)  based on similarity score and corresponding heatmaps (col. $4$). Firstly, the similarity heatmap of ProtoVerse is localised to the fracture regions of the vertebrae (col 2). 
   In contrast, post-hoc baselines fails to precisely localise clinically important regions in the vertebral body and focuses on the wrong vertebra.
   Secondly, the top $3$ prototypes visually explain why the highest activated region is important to grade the fracture type correctly. Note that, for all $3$ prototypes show positive class connections and belong to the same class.
   }
  \label{fig:sample_class-1}
\end{figure*}

\begin{figure}[!h]
    \centering
     \begin{subfigure}[b]{0.95\textwidth}
        \centering
        \includegraphics[width=\textwidth]{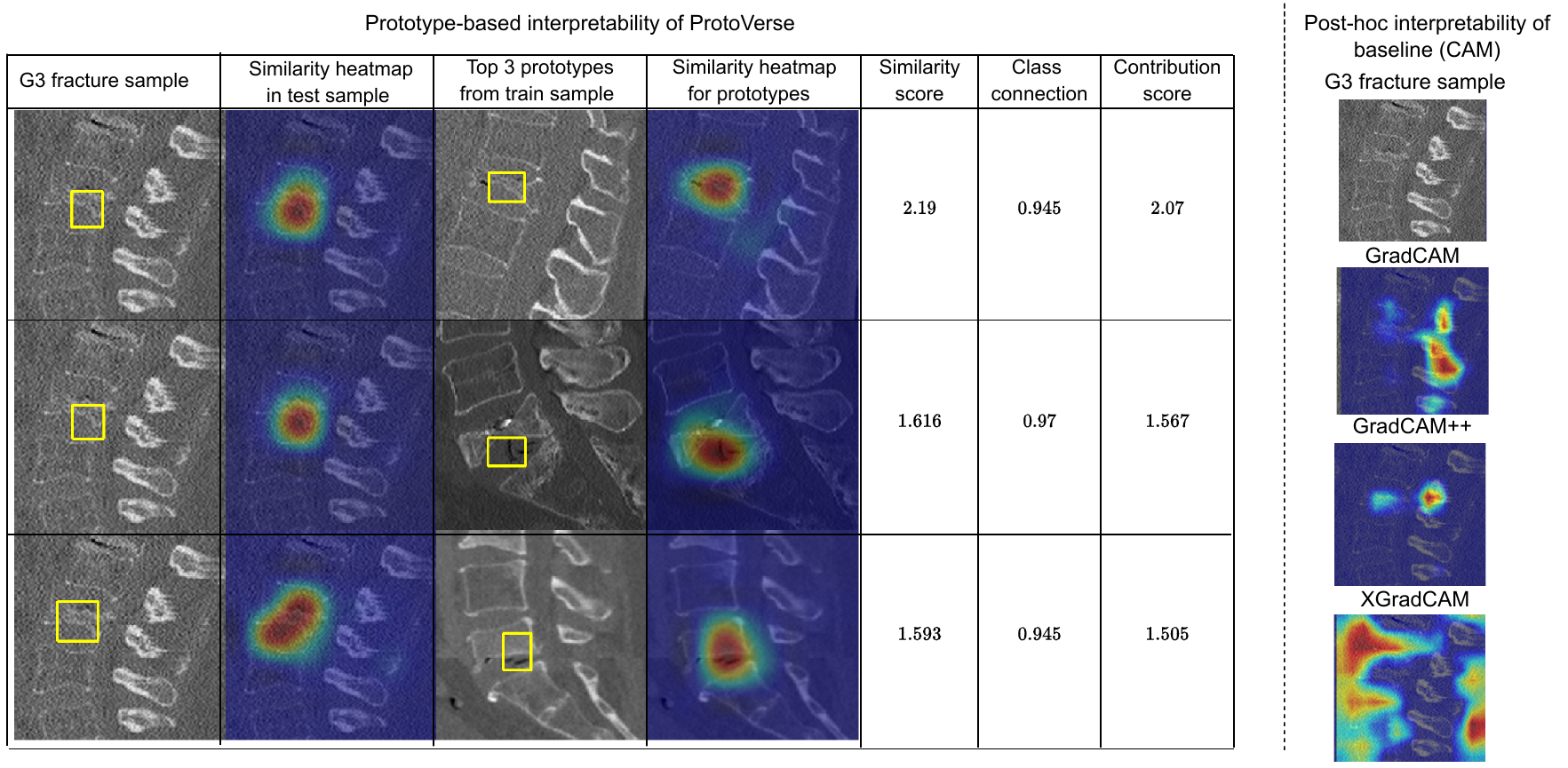}
    \end{subfigure}
    
    \vfill
    \begin{subfigure}[b]{0.95\textwidth}
        \centering
        \includegraphics[width=\textwidth]{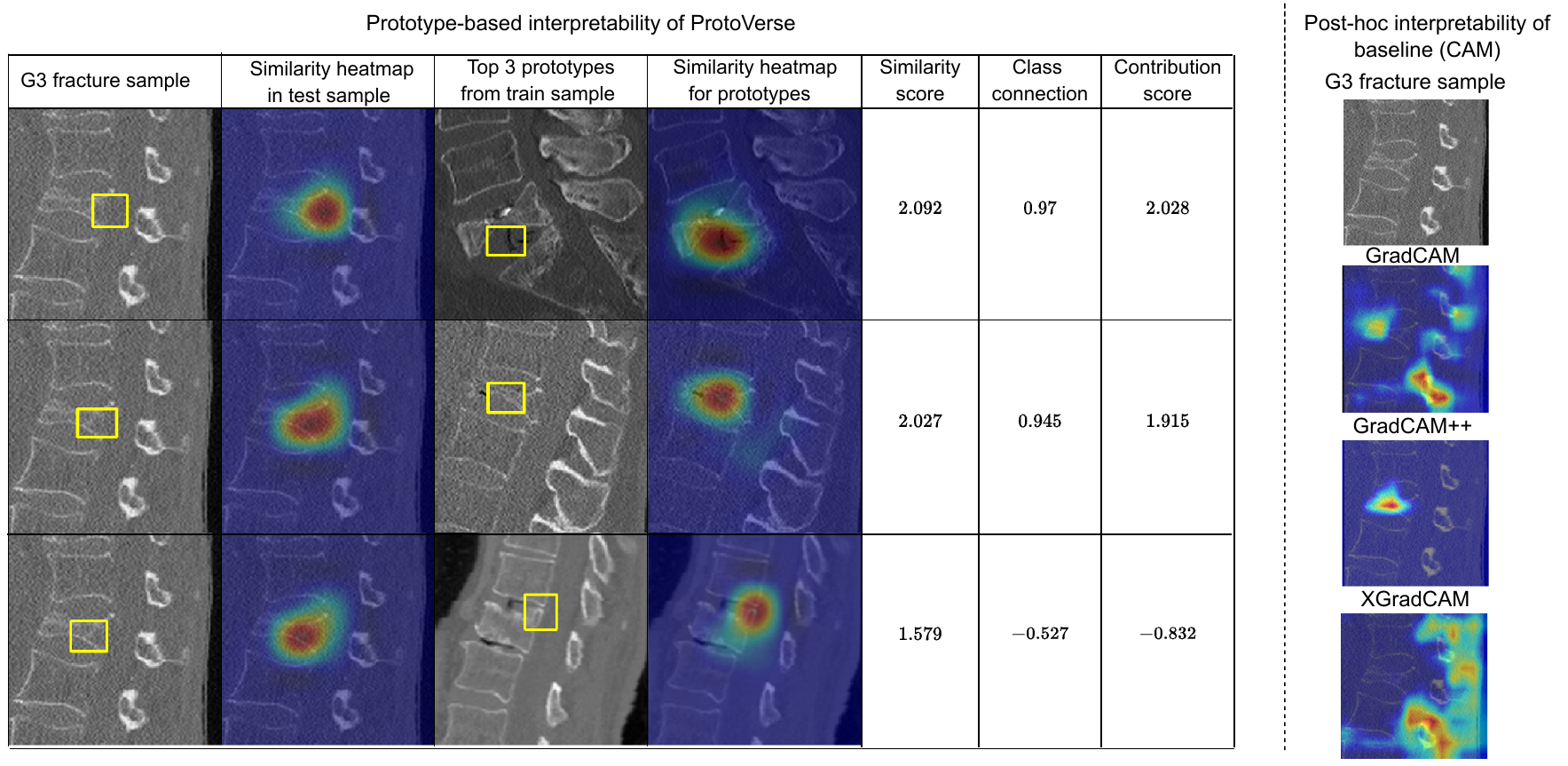}
    \end{subfigure}
      \vspace{-0.3em}
   \caption{Test-time interpretability of ProtoVerse (left) in comparison to post-hoc baselines (right) for two typical G3 fracture samples. 
   We show the top $3$ closest prototypes (col. $3$)  based on similarity score and corresponding heatmaps (col. $4$). Firstly, the similarity heatmap of ProtoVerse is localised to the fracture regions of the vertebrae (col 2). 
   In contrast, post-hoc baselines fails to precisely localise clinically important regions in the vertebral body and focuses on the vertebral processes.
   Secondly, the top $3$ prototypes visually explain why the highest activated region is important to grade the fracture type correctly. Note that, for first example, all $3$ prototypes show positive class connections and belong to the same class, whereas the third prototype for second example coming from G2 shows a negative class connection.
   }
   \label{fig:sample_class-2}
   
\end{figure}
We compare ProtoVerse's interpretability with the post-hoc methods on the baseline model using Class Activation Map (CAM) in Fig. \ref{fig:sample_class-1} and \ref{fig:sample_class-2} for representative test samples from G2 and G3, respectively. Here, we compute the \textit{similarity heatmap} (col. 2), \textit{similarity score}, and \textit{class connection score} as described in Section \ref{sec:method_learning}. The similarity score is multiplied by the class connection score to get the total contribution score of a given prototype to explain the classification output of a sample. We show the top $3$ most activated prototypes corresponding to each of the test samples and the corresponding heatmap in col. 3 and col. 4, respectively. Col. 1 shows the highest activated region in the similarity heatmap for the test samples corresponding to each of the top-$3$ prototypes. 

First sample in Fig. \ref{fig:sample_class-1} has the first prototype focuses from top-right to mid-bottom regions of the vertebra, while the last two prototypes focuses on the mid-right sections. 
Second sample in Fig. \ref{fig:sample_class-1} has the first two prototypes activating at the top-right corner of the vertebral body, while the last prototype attends to its middle section. The similarity heatmaps focus on distinct relevant contexts around the deformities of G2. Positive class connection scores for all three prototypes from G2 confirm human-interpretable reasoning for classifying the test sample as G2. 

Similarly, in both examples from Fig. \ref{fig:sample_class-2}, we observe that all the prototypes from G3 correspond to severely diminished height in the middle section of the vertebrae. These prototypes show high activations in similarly deformed sections at the middle and middle-right of the vertebrae. This observation reaffirms the reasoning ability of ProtoVerse based on meaningfully retrieved visual prototypes, e.g., similarly deformed vertebral edges. We also note that the last prototype (from G2) has a negative class connection since it does not belong to the class of the test sample despite having high visual similarity. Thus, the overall contribution score determines the role of the prototype in class prediction of the given sample. From both cases, one can observe the benefit of a diverse set of learnt prototypes in identifying different sub-parts of the fractured vertebrae of interest bringing in interpretability in the overall classification decision. 

For post-hoc CAM methods, the heatmaps are generated a posteriori using the class activations from the baseline model. In our experiments, we considered three popular variants of CAM based methods, namely GradCAM (\citep{selvaraju2017grad}), GradCAM++ (\citep{Chattopadhyay2017GradCAMGG} and XGradCAM (\citep{Fu2020AxiombasedGT}).

For the G2 samples, we observe that the most-activated (in red) is not highlighting the vertebrae of interest (center vertebrae) but focusing on the one above. For the G3 samples, the activation is scattered and does not focus on any vertebrae body for most of the methods except GradCAM++ for the second sample. Although the baseline model does not perform too badly in terms of classification accuracy, both cases indicate that it is barely interpretable post-hoc. This is because GradCAM variants only provides local interpretation based on sample-specific activation and ignores the recurring features important to explain the entire class, unlike the prototypes in ProtoVerse.



\subsection{Similarity-based Interpretability across Classes}
\label{sec:correlation_results}

\begin{table}[]
\centering
\begin{tabular}{@{}llll@{}}
\toprule
& \multicolumn{3}{c}{\textbf{Prototype Class}} \\
\textbf{Test Class} & \textbf{Healthy} & \textbf{G2} & \textbf{G3} \\ \midrule
Healthy                                                                                                     & 4.72             & -0.33            & -0.19            \\
G2                                                                                                     & 0.73             & 2.72             & -0.01            \\
G3                                                                                                     & 0.35             & -0.06            & 2.70             \\ \bottomrule
\end{tabular}
\caption{The average contribution score for all samples from each class has been correlated with
the prototypes from each class. Strong diagonal correlation indicates prototypes of each class are most activated and thus reliably explain samples from their class. }
\label{tab:correlation}
\end{table}
Here, we analyse how much one class's prototypes influence another class's decision-making process. For that, we average total contribution scores from each class's prototypes for all test samples of each class and present them in Table \ref{tab:correlation}.
We demonstrate that the average contribution for all the samples corresponding to the prototypes of their respective class is maximum (strong correlation along diagonal), thus confirming that the learnt prototypes in each class bear a strong correlation with the samples from the respective class. Additionally, the positive self-correlation of the healthy class ($4.72$) compared to negative cross-correlation in fracture classes G2, G3 ($-0.33,-0.19$) shows the model's strong anomaly detection (healthy vs fracture) performance.
To the best of our knowledge, our study is the first of its kind to successfully learn fine-grained prototypes to further distinguish between fracture grades beyond contemporary anomaly detection works \citep{mohammadjafariUsingProtoPNetInterpretable2021, weiMProtoNetCaseBasedInterpretable2023}.

\subsection{Clinical Validation Results} 
\label{sec: clinical_results}

Along with experimental results, we designed an expert 
clinical validation performed by two radiologists with $5$ mean years of experience. For this, we emphasised the clinical usability of the learnt prototypes in explaining a test image. 
We prepared a study including all test samples of clinically relevant G2 and G3 fracture classes.
For each sample, we listed the top $3$ closest prototypes from the training set based on the similarity score and asked the radiologists to rate based on the following two (Yes/No) questions (see Suppl. for details):
\begin{enumerate}
    \item  Are each of the retrieved prototypes relevant in explaining fracture grading of the test image, i.e., \textit{Prototype Relevance}?
    \vspace{-0.5em}
    \item Are the highest activated parts of the test image visually similar to the highest activated parts of the retrieved prototypes, i.e., \textit{Visual Similarity}?
\end{enumerate}



\begin{table}[h]
\centering
\begin{tabular}{l|lll|lll}
\toprule
Expert & \multicolumn{3}{c|}{\textbf{Prototype Relevance}} & \multicolumn{3}{c}{\textbf{Visual Similarity}} \\
Rater & G2 & G3 & Total & G2 & G3 & Total\\
\midrule
Rater-1      & $1.00$ & $1.00$ & $1.00$  & $0.33$ &  $0.72$ & $0.59$ \\
Rater-2      & $0.88$ & $0.78$ & $0.84$ & $0.61$ & $0.71$ & $0.65$ \\ 
\bottomrule
\end{tabular}

\caption{Clinical Validation results show the high relevance of the learnt prototypes from both experts in explaining the test-time decision. Doctors found relatively higher visual similarity for G3 than G2, indicating difficulty in learning G2 from small datasets.} 
\label{tab:clinical}
\end{table}

For both questions, we consider \lq Yes\rq   answer for a sample if $2$ out of $3$ prototypes are marked \lq Yes\rq  by the doctors. We report the percentage of \lq Yes\rq outcome for all questions in Tab. \ref{tab:clinical}. We observe that both radiologists find the retrieved prototypes to be relevant for adequately explaining the fracture class. G3 prototypes have a better average relevance score. 
We attribute this to the fact that G2 samples are not severely fractured and can often be semantically difficult to learn.
Qualitatively, the doctors have found that the prototype's activation patterns were more pronounced in the posterior elements of the vertebral body. This finding might be due to the algorithm's proficiency in recognizing higher-grade fractures, which usually involve the posterior aspect of the vertebral body, compared to low-grade fractures of the anterior vertebral body. Radiologists rate consistently high visual similarity for G3, further validating the visual explanation. However, for G2, inter-rater variability is more pronounced, which is commonly observed for lower fracture grades \citep{buckens2013intra}. Our future work will explore explicitly modeling inter-rater variability in the prototype layer to better learn a set of prototypes for small datasets.
\vspace{-0.3em}


\subsection{Quantitative Ablation Study} 
\label{sec:ablation}

In this section, we provide 5-fold cross-validation on the training set for hyper-parameter search and ablation on three aspects of our proposed ProtoVerse model.
First, we study the effect of the number of prototypes per class. Thirdly, we compare our median-weighted class imbalance loss with different loss weighting strategies. Finally, we show ablation on the coefficient of the Diversity loss.

\subsubsection{Ablation on Number of Prototypes}
Table \ref{tab:ablation_prototypes} reports quantitative performance using varying numbers of prototypes per class for ProtoVerse with VGG11 backbone. We observe that three prototypes per class are optimal. Since a single prototype per class severely hurts the interpretability of the model, we refrain from including it in the experiment.

\begin{table*}[!ht]
\centering
\begin{tabular}{@{}llll@{}}
\toprule
\textbf{\# Prototypes} & \textbf{Class-avg} & \textbf{Class-avg} & \textbf{Sample} \\
 & \textbf{Accuracy} & \textbf{F1} & \textbf{Accuracy} \\ \midrule
2                      & 72.24 $\pm$  7.60                      & 66.22 $\pm$ 9.80                    & 88.99 $\pm$ 4.87                  \\
3                      & 76.36 $\pm$ 5.95                       & 75.97 $\pm$ 4.80                    & 93.58 $\pm$ 1.37                    \\
4                      & 75.40 $\pm$ 3.06                       & 70.07 $\pm$ 5.82                    & 91.01 $\pm$ 3.53                    \\
5                      & 74.67 $\pm$ 7.33                       & 66.13 $\pm$ 11.28                   & 87.34 $\pm$ 7.35                    \\
 \bottomrule
\end{tabular}
\caption{
Performance comparison by varying the number of prototypes per class
}
\label{tab:ablation_prototypes}
\end{table*}

\begin{table*}[!ht]
\centering
\begin{tabular}{@{}llll@{}}
\toprule
\textbf{$\lambda_{\mbox{div}}$} & \textbf{Class-avg} & \textbf{Class-avg} & \textbf{Sample } \\
 & \textbf{Accuracy} & \textbf{F1} & \textbf{Accuracy} \\ \midrule
0.1 & 75.66 $\pm$ 3.89 & 69.45 $\pm$ 4.04 & 91.10 $\pm$ 1.10 \\
0.3 & 76.36 $\pm$ 5.96 & 75.97 $\pm$ 4.80 & 93.58 $\pm$ 1.37\\
0.5 & 75.12 $\pm$ 2.10 & 71.99 $\pm$ 4.58 & 92.84 $\pm$ 2.06 \\ 
\bottomrule
\end{tabular}
\caption{
Performance comparison for different weighting of Diversity loss.
}
\label{tab:ablation_lambda}
\end{table*}

\subsubsection{Ablation on Weighting Strategy}

\begin{figure}[ht]
    \centering
     \includegraphics[width=0.5\linewidth]{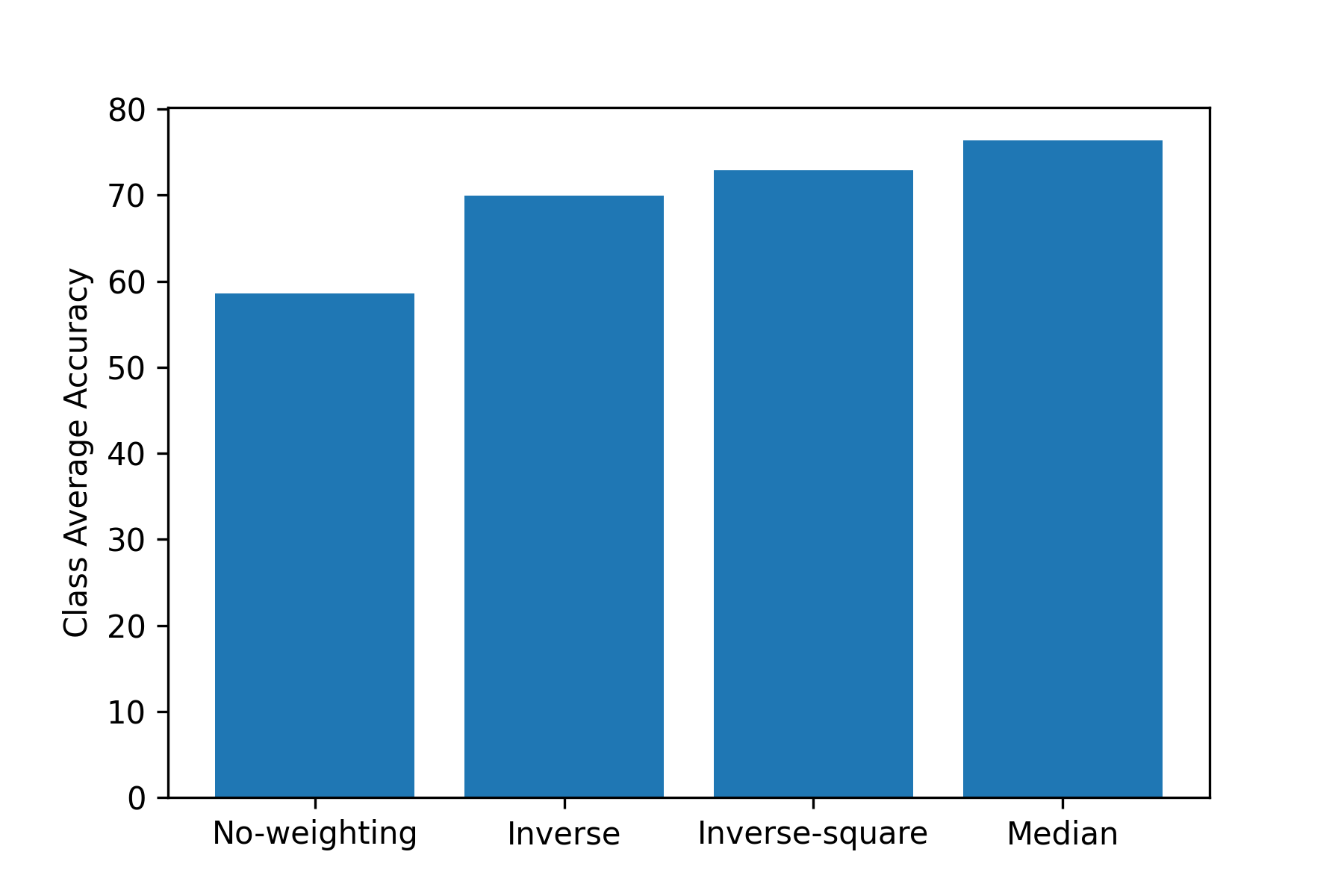}
   \caption{
   Bar plot showing comparison among different class weighting strategies
   }
   \label{fig:class_weight}
\end{figure}

Fig. \ref{fig:class_weight} shows ablation on different class weighting strategies to mitigate the class imbalance issue for VCF grading using three prototypes per class. Note that among all competing weights, median weighting produces the best result.

\subsubsection{Ablation on Diversity Loss Coefficient}

Table \ref{tab:ablation_lambda} shows the ablation of coefficient for Diversity loss. We use the above optimal parameters for the model. We note that $0.3$ is a suitable choice for sufficient variation in the visual prototype without affecting the model's classification performance. 
\subsection*{Limitation}

We believe improving the dataset imbalance with an emphasis on diverse fracture grade samples (potentially G1 fractures) could further improve the quality of the prototypes and clinical utility of such interpretable models. Lastly, although the learnt prototypes are designed to be human-understandable, nonetheless some prototypes are not human-interpretable and it still remains to be explored whether the model reasoning is incorrect in such cases or the generated explanation needs to be enhanced to be better human-aligned.

\section{Conclusion}
In this work, we proposed a novel IBD method, ProtoVerse, for VCF grading, leveraging prototype-based representation learning. Specifically, we address the challenge of learning diverse yet representative prototypes from a small dataset with a high class-imbalance. The resultant ProtoVerse outperforms previous IBD models, such as ProtoPNet, and offers enhanced interpretability that non-IBD baselines. We have validated our results with radiologists, confirming the clinical applicability of the proposed method.
To further enhance the quality of learnt prototypes, future research could include human-in-loop feedback during training to identify and replace unsatisfactory prototypes.

\acks{This work was funded by the Bavarian Ministry for Economic Affairs, Regional Development and Energy as part of a project to support the thematic development of the Institute for Cognitive Systems (IKS).
SS and JSK are supported by European Research Council (ERC) under the European Union’s Horizon 2020 research and innovation program (101045128-iBack-epic-ERC2021-COG).}

%
\ethics{The work follows appropriate ethical standards in conducting research and writing the manuscript, following all applicable laws and regulations regarding treatment of animals or human subjects.}

\coi{We declare we don't have conflicts of interest.}

\bibliography{main.bib}


\clearpage
\appendix
\section{Details on Clinical Validation}

\begin{figure}[!ht]
    \centering
     \includegraphics[width=0.95\textwidth]{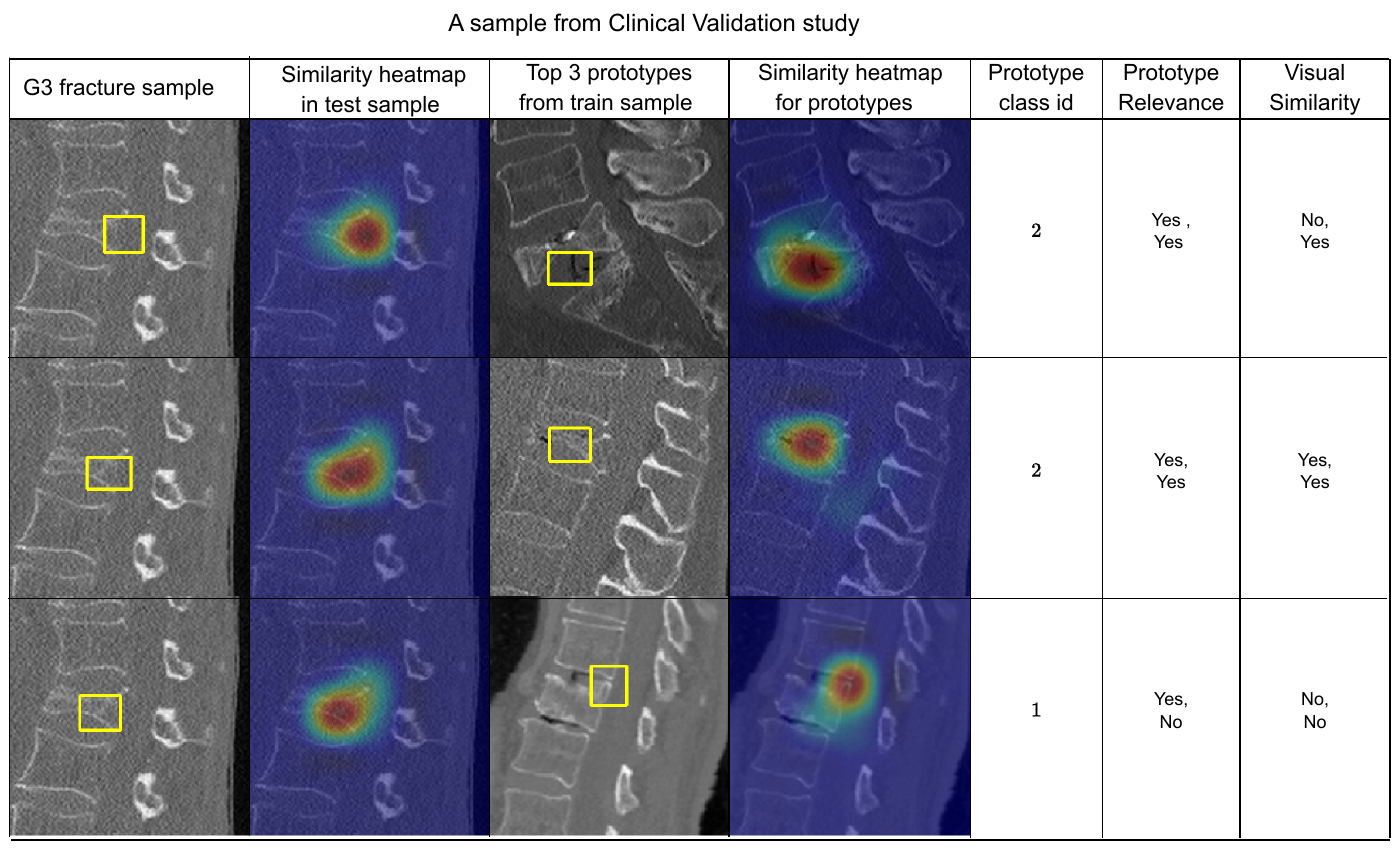}
   \caption{Exemplary samples from clinical evaluation done by radiologists. The answers obtained from the two radiologists are indicated.}
   \label{fig:clinical_eval_sample}
\end{figure}

Fig. \ref{fig:clinical_eval_sample} shows a sample from the clinical validation report. The radiologists are provided with the top-3 activated prototypes along with the highest-activated regions in the test image. Additionally, they are given the similarity heatmap of each of the top-3 prototypes for visual reasoning. Both radiologists provided \lq yes/no \rq answers to the questions and overall opinion on the findings.

\end{document}